\RequirePackage[OT1]{fontenc}
\documentclass[letterpaper,10pt,conference]{ieeeconf}
\IEEEoverridecommandlockouts
\usepackage{amsmath,amssymb}
\usepackage{graphicx}
\usepackage{booktabs,array,tabularx}
\usepackage{xcolor}
\usepackage{tikz}
\usetikzlibrary{arrows.meta,positioning,calc}
\usepackage{cite}
\usepackage{url}
\usepackage{microtype}
\definecolor{traincolor}{RGB}{232,243,250}
\definecolor{freezecolor}{RGB}{239,241,244}
\newcommand{\method}{ReLAP}

\title{\LARGE\bf ReLAP: Latent Action Pretraining through\protect\\ Recurrent  Video Prediction in JEPA Space}
\author{R. Gnana Praveen, Ali Rouzbayani, Zhiying Jiang, Bidhan Roy\protect\\ Bagel Labs}
\begin{document}
\maketitle
\thispagestyle{empty}
\pagestyle{empty}

\begin{abstract}
Videos provide a rich source of physical interaction for learning robot behavior without action annotations, yet effective latent actions must capture not only local transitions but also how those transitions evolve over time. We introduce \textbf{ReLAP}, a latent-action pretraining framework that couples local transition learning with recurrent multi-step prediction in a frozen JEPA representation space. A shared inverse dynamics model infers compact continuous latent actions from adjacent observations, while a latent world model recurrently predicts the future JEPA features. Because each predicted feature becomes the context for the next transition, errors at later steps backpropagate through the rollout to earlier latent actions, encouraging temporally coherent dynamics beyond pairwise reconstruction. 
We transfer this representation to a vision-language model using two complementary objectives: latent distillation from the frozen inverse teacher and recurrent future-feature prediction through the frozen latent world model. The resulting policy is then adapted to robot demonstrations with action flow matching. At deployment, the explicit latent-action and world-model branches are removed, and the fine-tuned VLM predicts physical actions directly, reducing inference complexity.
With only 11M pretraining samples and 1.06B trainable parameters, ReLAP achieves \(97.9\%\) on standard LIBERO and \(70.0\%\) over zero-shot LIBERO-Plus perturbations. On the LARY latent-action benchmark, ReLAP ranks second in both mean regression error and mean classification accuracy among the compared methods. These results demonstrate that recurrent action-free latent action pretraining can learn compact, temporally structured, and transferable representations for downstream robot control.
\end{abstract}

\section{Introduction}

Videos of human and robot interaction provide abundant supervision for
robot learning without requiring physical action annotations. Latent action
models exploit such action-free data by inferring compact pseudo-actions
from visual transitions. A typical latent action model combines an inverse
dynamics model, which infers a latent action from two consecutive
observations, with a forward dynamics model that reconstructs the resulting
transition. These inferred actions can then serve as supervision for
pretraining vision-language-action policies before adaptation to
action-labeled robot demonstrations.
While effective, pairwise transition reconstruction does not explicitly
require consecutive latent actions to remain compatible over time. Each
transition is typically reconstructed from its observed starting state, so
accurate one-step predictions do not necessarily yield an accurate
multi-step trajectory when predicted states become the context for later
transitions. This distinction is particularly important for manipulation:
a reach changes the state from which a grasp begins, and the grasp changes
the state from which a lift is executed.
Recent work addresses this temporal structure in different ways. RotVLA
introduces structured continuous latent actions, while ALAM and related
compositional latent-action models impose explicit consistency constraints
on how latent transitions combine. Other approaches use future prediction
or world modeling to enrich latent-action supervision. In contrast, we
encourage temporal coherence through the predicted consequences of the
latent sequence itself, without prescribing an explicit algebraic
composition rule in latent space.
Specifically, we recurrently roll out the inferred latent actions in a
frozen JEPA representation space, where each predicted feature becomes the
input to the next transition. This allows errors at later horizons to
propagate through the predicted trajectory and supervise earlier latent
actions. Combined with local transition reconstruction, this objective
preserves the grounding of each latent to its observed transition while
encouraging the full sequence to remain temporally coherent under
multi-step prediction.

\begin{figure}[!t]
\centering
\includegraphics[width=0.4\textwidth]{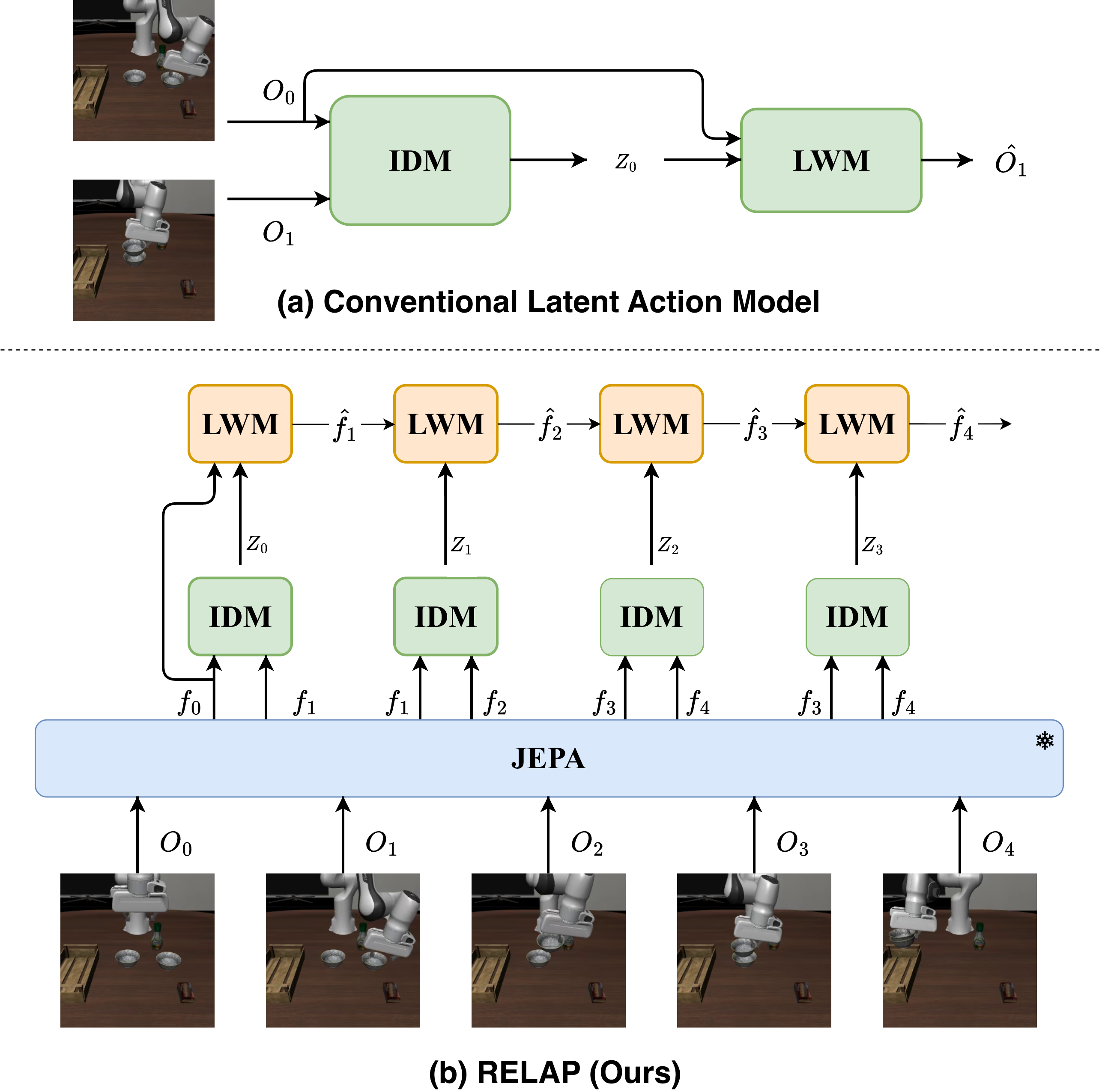}
\caption{(a) Conventional latent action models infer a latent action from an observed
transition and reconstruct its immediate consequence.
(b) ReLAP infers a sequence of continuous latent actions and recurrently
rolls them out in JEPA feature space, allowing later prediction errors to
supervise earlier latent actions and encouraging temporal coherence across
the sequence.}
\label{fig:teaser}
\end{figure}

We introduce \textbf{ReLAP}, a framework for latent-action pretraining through
recurrent video prediction in JEPA space (Fig.~1). A shared inverse dynamics
model infers a continuous latent action from each adjacent observation pair
encoded by a frozen V-JEPA~2.1 encoder~\cite{vjepa21}. Starting from the first
observed feature map, a shared forward dynamics model recurrently applies the
inferred latent sequence, using each predicted feature as the input to the next
transition. We supervise every rollout step and backpropagate through the
predicted intermediate states, allowing later prediction errors to constrain
earlier latent actions. A local objective anchors each latent to its observed
transition, while recurrent prediction encourages the sequence to remain
temporally coherent under its own predicted dynamics.
In Stage~2, a vision-language model predicts the complete latent sequence from the
current image and language instruction in a single forward pass. The inverse
and forward models remain frozen: latent distillation aligns the VLM predictions
with the video-inferred actions, while recurrent future-feature supervision
evaluates their cumulative effects under the same learned dynamics. In Stage~3,
robot demonstrations introduce physical-action supervision through flow
matching, transferring the pretrained transition knowledge to executable
control. The latent rollout spans \(1.6\,\mathrm{s}\), while the policy executes
a \(0.4\,\mathrm{s}\) action chunk before replanning. Future observations and
the inverse teacher are used only during training; at deployment, the explicit
latent-action and JEPA dynamics branches are discarded, and the fine-tuned VLM
directly predicts physical actions.
We evaluate the downstream control and rigorously analyze the learned latent action
representations. ReLAP achieves \(97.9\%\) success over 2,000 episodes on standard
LIBERO~\cite{libero} and \(70.0\%\) pooled success over 10,030 zero-shot
LIBERO-Plus perturbations~\cite{liberoplus}. On the LARY
benchmark~\cite{lary}, ReLAP obtains \(0.24\) mean regression MSE and
\(65.76\%\) mean classification accuracy, ranking second among the SOTA
latent-action models. Targeted representation analyses further show that the
learned latents are sample-specific, temporally structured, and aligned with
corresponding physical actions.
Our contributions are threefold:
\begin{itemize}

\item \textbf{Recurrent latent-action learning in JEPA space.}
We couple local transition reconstruction with recurrent prediction in frozen
V-JEPA~2.1 features, allowing later rollout errors to supervise earlier
video-inferred latent actions through predicted intermediate states.

\item \textbf{Dynamics-aware transfer to a VLM.}
We combine latent distillation with recurrent future-feature supervision to transfer the discovered latent actions in stage 1 to VLM pretraining and finetuning, thereby leveraging the benefits of both inverse dynamics model and forward model of stage 1.

\item \textbf{Comprehensive representation and control evaluation.}
We evaluate ReLAP on LIBERO, zero-shot LIBERO-Plus, and LARY, together with
targeted probes of sample specificity, temporal structure, and physical-action
alignment.

\end{itemize}

\section{Related Work}

\subsection{Latent Action Learning from Action-Free Video}

Latent action models (LAMs) provide a scalable way to learn action-like
representations from videos without physical action annotations. Early
approaches such as LAPA learn discrete latent actions through inverse--forward
reconstruction and use them as pseudo-action targets for VLA
pretraining~\cite{lapa,lapo,genie}. Subsequent work explored richer
representations by exploring task-centric discrete actions in DINO feature
space~\cite{univla}, while CoMo \cite{como} and Garrido et al.\cite{jepalam} show the benefits of
continuous latent actions for capturing fine-grained motion and diverse
in-the-wild dynamics. JALA~\cite{jala} instead avoids explicit
observation reconstruction and aligns predictive action embeddings with
inverse dynamics and physical actions.
Most of these approaches primarily optimize the quality of individual
transitions or the predictability of latent actions. ReLAP instead
focuses on whether a sequence of video-inferred actions remains
predictive when its own earlier consequences become the context for later
transitions. This introduces recurrent supervision directly into latent-action
discovery rather than only into the downstream policy.

\subsection{Temporal Structure and Future Prediction}

Recent works highlight that independently reconstructed latent
actions need not form a coherent sequence. Moto~\cite{moto} models latent motion tokens
autoregressively, while ViPRA jointly predicts future visual
observations and fine-grained latent-action sequences and regularizes the
underlying motion representation with perceptual and optical-flow
objectives~\cite{vipra}. Other approaches impose explicit structure directly
in the latent space. AC-LAM~\cite{wei2026learningadditivelycompositionallatent} enforces additive composition and inverse
consistency; RotVLA~\cite{rotvla} represents actions on a rotational manifold
and uses frame triplets to capture temporal relations; and ALAM~\cite{alam}
and DLAM~\cite{dlam} impose composition and reversal constraints on deterministic and
distributional latent transitions, respectively.
These approaches establish the importance of temporal structure, but largely
encode it either through explicit algebraic constraints.
ReLAP takes a complementary view: temporal compatibility is defined by the
\emph{predicted consequences} of the latent sequence. Each action is applied
to the feature state produced by the preceding action, and losses at later
horizons backpropagate through these predicted states to earlier inferred
actions. Thus, temporal structure emerges through learned state-dependent
dynamics without adding additional complexities of additive, rotational, or other fixed
composition rule in latent space.

Future prediction provides another route for strengthening latent-action
learning. LAWM~\cite{lawm} uses world modeling as supervision for latent-action
pretraining, while CoLA-World~\cite{colaworld} and CLAW~\cite{claw} jointly optimize latent
actions with expressive world models, allowing predictive losses to shape the
learned action interface. ViPRA~\cite{vipra} similarly combines
latent-action prediction with future visual prediction. ReLAP is
closest in spirit to this line of work, but differs in where the temporal
constraint is applied: rather than evaluating future prediction only from
observed context, it recurrently feeds predicted intermediate states back into
the forward model, explicitly coupling earlier latent actions to the accuracy
of later transitions.

\subsection{Feature-Space Dynamics for Robot Learning}

Predicting in pretrained visual feature spaces avoids the cost and ambiguity
of reconstructing every pixel. DINO-WM~\cite{dinowm} predicts future DINO features
conditioned on physical actions for test-time planning, while
V-JEPA~2-AC~\cite{vjepa2} learns action-conditioned dynamics in V-JEPA representations from
robot interaction data. In latent-action learning, UniVLA~\cite{univla} moves
reconstruction from pixels to DINO features. Most closely
related to ReLAP, Garrido et al.~\cite{jepalam} jointly learn a continuous inverse model and
world model in frozen V-JEPA~2-L features from action-free video. Their forward model, however, is trained with teacher forcing,
so each next-state prediction is conditioned on observed history rather than
the model's recursively predicted intermediate states.
JEPA representations are also increasingly used to provide predictive
supervision directly to robot policies. VLA-JEPA~\cite{vlajepa} predicts future latent states
without pixel generation, while JEPA-WAM~\cite{jepawam} couples transition
prediction and continuous action generation in pretrained V-JEPA
space. FLARE~\cite{flare} similarly aligns policy representations with
future latent observations. These methods demonstrate the value
of predictive feature-space supervision, but do not address the same problem
as ReLAP: learning a sequence of latent actions from action-free video by
allowing future rollout errors to shape the inverse model that inferred the
earlier actions.

ReLAP lies at the intersection of three emerging directions: label-free
latent-action discovery, temporally structured latent actions, and predictive
world modeling in pretrained feature space. Prior work has explored each
component separately---including future prediction, explicit latent
composition, and JEPA-space dynamics---but their intersection remains less
studied. ReLAP combines them through a simple recurrent principle: local
transition reconstruction grounds each latent action, while multi-step rollout
through predicted JEPA states couples the sequence through its decoded visual
effects. This provides temporal supervision without physical action labels,
pixel reconstruction, or an explicit algebraic composition rule.

\section{Proposed Approach}
\label{sec:method}


\begin{figure*}[!t]
\centering
\includegraphics[width=0.78\textwidth]{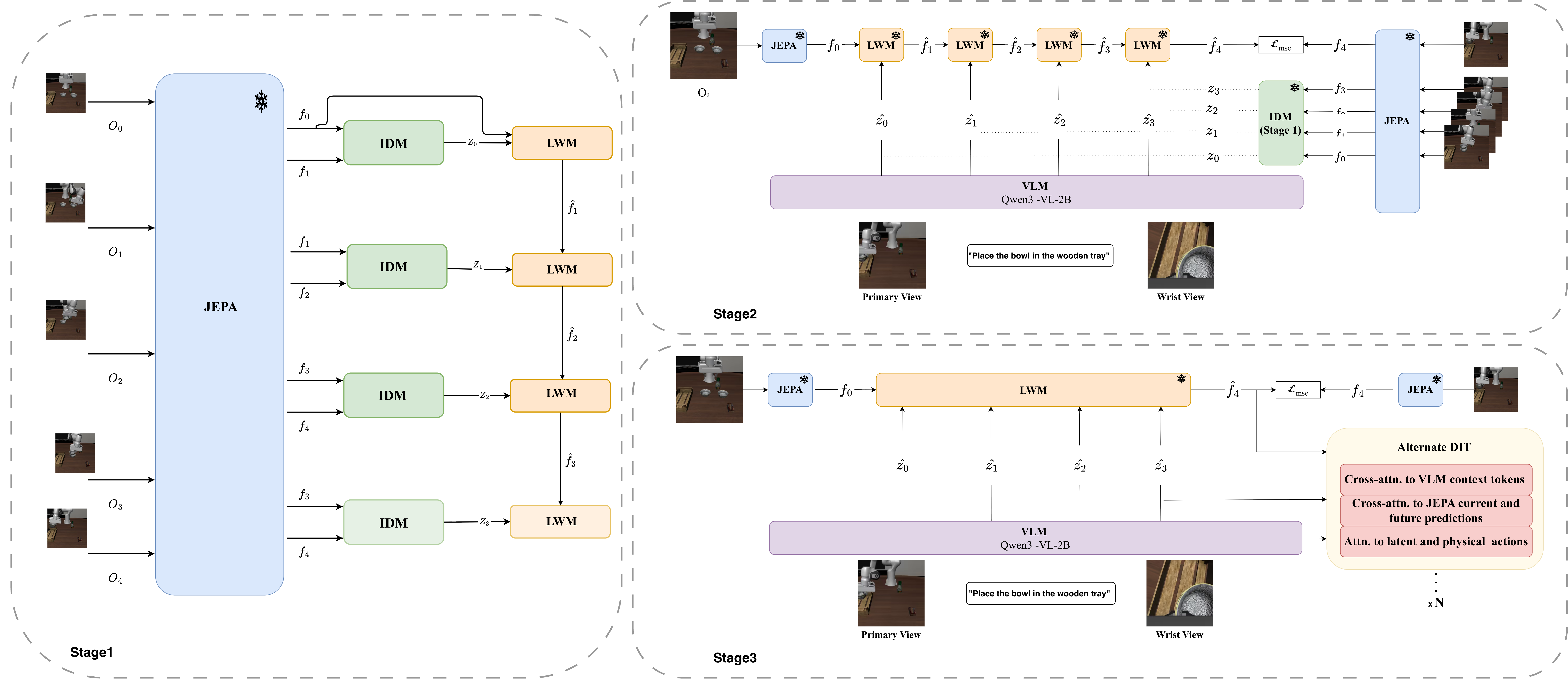}
\caption{Overview of \method{}. Stage~1 learns temporally coherent latent actions from unlabeled human and robot video through local and recurrent prediction in frozen JEPA features. Stage~2 trains a vision-language model to predict the latent sequence from the current image and instruction under the frozen Stage-1 models. Stage~3 adds a flow-matching action expert of alternating DiT blocks conditioned on the VLM context, the predicted latent actions, and the current and predicted future JEPA features; the forward model stays trainable under the terminal feature loss.}
\label{fig:architecture}
\end{figure*}


\method{} learns latent actions from action-free video by requiring their predicted latent actions to remain coherent over a sequence. The central mechanism is a shared latent world model that consumes its own previous feature prediction at each step. 
Figure~\ref{fig:architecture} summarizes the three stages: learning this representation from video, predicting it from the current observation and instruction, and adapting it to robot control. Physical action labels are used only in the third stage.

\subsection{Stage 1: Learning Latent Actions through Recurrent Prediction in JEPA Space}
\label{sec:stage1}

Stage~1 learns compact latent actions that explain local visual transitions
while remaining consistent over a multi-step rollout. Given a sequence of
\(H{+}1\) observations \(o_{0:H}\), we extract patch features
\(f_t=E(o_t)\in\mathbb{R}^{N\times d_f}\) using a frozen
V-JEPA~2.1 encoder~\cite{vjepa21}. We use \(H=4\) transitions sampled at
\(0.4\,\mathrm{s}\) intervals and perform all prediction directly in the
JEPA representation space.
For each adjacent feature pair, the inverse model \(I_\phi\) infers a
\(d_z=16\)-dimensional continuous latent action. 
Each latent is inferred independently from its corresponding transition.
A low-dimensional bottleneck together with KL regularization toward
\(\mathcal{N}(0,I)\) encourages a compact representation of the observed
visual change. The forward latent world model \(F_\psi\) predicts how a latent action
transforms the current feature map. We first optimize a
\emph{local transition loss}, which reconstructs each observed transition
independently by minimizing the Smooth-L1 distance between
\(F_\psi(f_t,z_t)\) and \(f_{t+1}\), with \(z_t\) sampled from the posterior.
This anchors each latent action to its corresponding local transition. Local reconstruction alone, however, does not require consecutive latent
actions to remain compatible when applied sequentially. We therefore
introduce recurrent prediction. Starting from the first observed feature,
the model recursively applies the inferred latent sequence,
\begin{equation}
\hat f_0=f_0,\qquad
\hat f_{t+1}=F_\psi(\hat f_t,\mu_t),
\quad t=0,\ldots,H-1 .
\label{eq:rollout}
\end{equation}
The rollout loss \(\mathcal{L}_{\mathrm{roll}}\) measures the Smooth-L1
distance between each predicted feature \(\hat f_{t+1}\) and its
corresponding target \(f_{t+1}\). Gradients are propagated through the
entire rollout, allowing later prediction errors to supervise earlier
latent actions through the intermediate predicted states. Consequently,
the latent sequence is encouraged to capture transitions whose effects
remain predictive as they become the context for subsequent transitions,
without imposing an explicit composition rule in latent space.
The complete Stage~1 objective is
\begin{equation}
\mathcal{L}_{1}
=
\mathcal{L}_{\mathrm{local}}
+
\lambda_r \mathcal{L}_{\mathrm{roll}}
+
\beta \mathcal{L}_{\mathrm{KL}},
\label{eq:stage1}
\end{equation}
where
\(\mathcal{L}_{\mathrm{KL}}\) denotes the average KL divergence between
the latent posterior and \(\mathcal{N}(0,I)\). We gradually introduce
\(\lambda_r\) after an initial period of local transition learning and
ramp \(\beta\) early in training. The visual encoder remains frozen
throughout Stage~1.
Because the inverse model observes both sides of each transition,
Stage~1 learns a representation of observed visual dynamics rather than
predicting future actions from the current observation. 

\label{sec:stage1}

\subsection{Stage 2: Latent Action Pretraining with VLM}
\label{sec:stage2}

Stage~2 trains a vision-language model (VLM) to predict the latent sequence from the current main-camera image $o_0$ and instruction $s$. We denote the resulting student predictor by $P_\theta(o_0,s)=(\tilde z_0,\ldots,\tilde z_{H-1})$. A Q-Former head uses \(H\) learned latent queries to attend to the VLM hidden
states and predicts all latent actions jointly in a single forward pass.
Unlike autoregressive decoding, the latent queries use bidirectional
self-attention, allowing each slot to interact with the others and encouraging
the predicted latent actions to form a temporally coherent sequence. Future
observations are used only to construct training targets and are not available
to the student input.
Stage~2 is supervised by two complementary objectives: \emph{latent
distillation} and \emph{future-feature prediction}. We freeze the visual
encoder \(E\), inverse model \(I_\phi\), and forward latent world model
\(F_\psi\). The posterior means
\(\mu_t\) produced by \(I_\phi\) serve as latent-action targets, while the
student-predicted sequence is recursively rolled out through the frozen forward
model:
\[
\tilde{f}_0=f_0,\qquad
\tilde{f}_{t+1}=F_\psi(\tilde{f}_t,\tilde{z}_t).
\]
The Stage~2 objective is
\begin{equation}
\mathcal{L}_{2}
=
\frac{\lambda_d}{H d_z}
\sum_{t=0}^{H-1}
\left\|
\tilde{z}_t-\mu_t
\right\|_2^2
+
\frac{\lambda_f}{H}
\sum_{t=0}^{H-1}
\ell_{\mathrm{JEPA}}
\left(
\tilde{f}_{t+1},f_{t+1}
\right),
\label{eq:stage2}
\end{equation}
where the first term aligns the VLM predictions with the Stage~1 latent-action
space, while the second evaluates whether those predictions reproduce the
corresponding future V-JEPA feature trajectory.
These two objectives play complementary roles. Latent distillation preserves
the semantic coordinates learned by the inverse model, whereas future-feature
prediction constrains the predicted latent sequence through its cumulative
visual effects. Although \(F_\psi\) remains frozen, gradients propagate through
the full recursive rollout to the Q-Former and VLM, encouraging latent actions
that are individually consistent with the teacher and jointly predictive of
future visual dynamics. The rollout loss is introduced gradually after initial
latent distillation to stabilize optimization.
Since the inverse teacher observes future frames, its posterior latents provide
training references rather than unique action targets: multiple latent
trajectories may be compatible with the same current observation and
instruction. Stage~2 therefore learns a future-free latent prior that predicts
a coherent transition sequence from the current visual-language context alone,
without using physical-action supervision.


\subsection{Stage 3: Adapting Latent Representations to Robot Control}
\label{sec:stage3}
Stage~3 adapts the pretrained representation to physical robot control using
action-labeled demonstrations. The policy receives the current main-camera
image, wrist image, and language instruction. The latent queries are positioned
before the wrist-view tokens in the causal VLM sequence, so latent prediction
depends only on the main view and instruction, while the action pathway can
additionally exploit wrist-view information. The Stage~2 Q-Former is retained,
with bidirectional attention among the four latent queries to preserve
sequence-level consistency.
During training, the predicted latent sequence is recurrently decoded by the
Stage~1 forward model to obtain its anticipated future JEPA representation.
A flow-matching action head then predicts an eight-step, seven-dimensional
action chunk from the VLM context. We additionally expose the action head to
the predicted latent sequence and current/future JEPA features during training,
providing complementary semantic and visual-dynamics context.
For a demonstration action chunk \(A\), we train the action head with standard
conditional flow matching. Using a noisy interpolation \(x_\tau\) between
Gaussian noise and \(A\), the model predicts the corresponding velocity
\(A-\epsilon\). The complete Stage~3 objective is

\begin{equation}
\mathcal{L}_{3}
=
\mathcal{L}_{\mathrm{flow}}
+
\lambda_d\mathcal{L}_{\mathrm{distill}}
+
\lambda_w\mathcal{L}_{\mathrm{world}},
\label{eq:stage3}
\end{equation}
where \(\mathcal{L}_{\mathrm{distill}}\) preserves alignment with the frozen
Stage~1 inverse teacher and \(\mathcal{L}_{\mathrm{world}}\) matches the
predicted terminal JEPA feature to the observed future feature. We use
\(\lambda_d=\lambda_w=0.1\). Thus, Stage~3 jointly optimizes executable action
prediction while retaining the transition structure acquired during
action-free pretraining.
The visual encoder and inverse teacher remain frozen, while the VLM, latent
prediction head, forward model, and action head are adapted to robot
demonstrations. Importantly, the latent-action and JEPA world-model branches
serve as training-time supervision rather than mandatory deployment modules.
At inference, we retain only the fine-tuned VLM direct-action pathway and
discard the latent decoder and visual-dynamics attention branches. Future
observations and the inverse teacher are therefore never required at
deployment, yielding a simpler and lower-latency policy.

\section{Experimental Setup}

\begin{figure}[t]
\centering
\includegraphics[width=\linewidth]{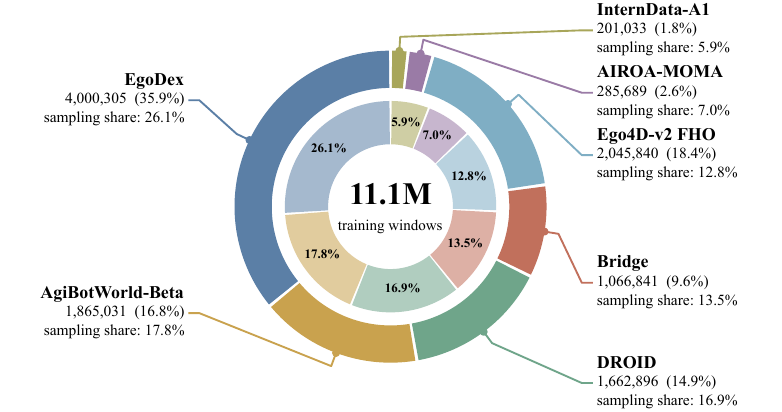}
\caption{Stage-1 video mixture. Outer ring: each source's share of the 11.1M available 1.6-second training windows. Inner ring: its share of sampled training examples under the mixture weights.}
\label{fig:datamix}
\end{figure}

\textbf{Datasets.} 
Stages~1 and 2 train on a mixture of seven large-scale video datasets: Bridge-v2~\cite{bridgev2}, DROID~\cite{droid}, AgiBot~\cite{agibotworld}, EgoDex~\cite{egodex}, Airoa-MOMA~\cite{airoamoma}, InternData~\cite{interndata}, and Ego4D~\cite{ego4d}. These datasets provide diverse demonstrations of physical interactions in both human and robot environments. In stage 1, we use only the raw video streams without accessing any robot action annotations, proprioceptive signals, or language instructions, ensuring that the latent actions learned are derived exclusively from visual dynamics. We follow the preprocessing strategy of OSCAR \cite{oscar} to process raw videos, standardizing frame rates and resolutions while preserving the temporal integrity of the interactions. Specifically, videos are sampled at 20 Hz, and each training clip consists of 5 frames spaced 0.4 seconds apart, yielding 11,127,635 training windows as shown in Fig~\ref{fig:datamix}. In stage 3, we use the LIBERO benchmark suite \cite{libero}, which includes 40 tasks across four suites: LIBERO-Spatial, LIBERO-Object, LIBERO-Goal, and LIBERO-10, resulting in 2K demonstrations and perform downstream evaluation of LIBERO and LIBERO plus benchmarks. 

\textbf{Implementation Details.}
ReLAP learns latent actions in the V-JEPA representation space. We use a frozen V-JEPA~2.1 ViT-B~\cite{vjepa21} encoder on $(256\times256)$ RGB frames, producing 256 visual tokens per frame. In Stage~1, the inverse dynamics model consists of 24 transformer layers and the forward latent world model uses six transformer layers, both with hidden width 1024. Each transition is represented by a 16-dimensional continuous latent action, and four consecutive transitions span a $1.6\,\mathrm{s}$ clip. Stage~1 is trained for 100k steps using AdamW with a learning rate of $3\times10^{-4}$ and an effective batch size of 512. The recurrent prediction loss weight $\lambda_r$ is gradually increased to 2, while the KL coefficient $\beta$ is ramped to \(5\times10^{-5}\) early in training for stable optimization. In Stage~2, we initialize from Qwen3-VL-2B and retain its first 16 text Transformer blocks. Eight learned latent-query tokens interact with the backbone through a single-layer Q-Former. Stage~2 is trained with an effective batch size of 256, with the recurrent prediction objective gradually introduced from 5k steps. For Stage~3, we initialize the latent-action model from the final Stage~1 checkpoint and the vision-language backbone from the 80k-step Stage~2 checkpoint, while initializing the action expert from scratch. The action expert is a 16-block flow-matching Transformer with hidden width 1024. The policy receives the current main-camera image, wrist-camera image, and language instruction; no proprioceptive state is used. In the causal input sequence, latent queries are placed before the wrist-view tokens, such that latent actions are predicted only from the main view and language instruction, while the action expert can additionally access wrist-view information through the full backbone hidden sequence. Stage~3 is optimized using AdamW with a learning rate of \(10^{-4}\), batch size 256, 1,500 warmup steps, and cosine decay to \(5\times10^{-7}\). Gradients are clipped to a maximum norm of 1. The loss weights are 1.0 for action flow matching, 0.1 for terminal future-feature prediction, and 0.1 for latent-action distillation. 
Throughout Stage~3, the visual encoder and inverse dynamics model remain frozen, while the forward latent world model, latent prior, action expert, and VLM backbone are optimized. 
\\
\textbf{Evaluation Benchmarks}
We evaluate on LIBERO [10] and LIBERO-Plus [11].
LIBERO has four suites, Spatial, Object, Goal, and Long (the
libero 10 suite), each with ten tasks. We run 50 rollouts
per task, giving 2,000 episodes. LIBERO-Plus perturbs the
same tasks along seven axes: camera viewpoint, robot ini-
tial state, language instruction, lighting, background texture,
sensor noise, and object layout. It contains 10,030 perturbed
task instances, and following the benchmark convention we
run each once. The same Stage-3 checkpoint is used on
both benchmarks with no fine-tuning on perturbed tasks, evaluating the zero-shot generalization to LIBERO plus.
Because the Stage-3 demonstrations cover the same 40 tasks
that LIBERO evaluates, standard LIBERO measures down-
stream task performance, while LIBERO-Plus measures how far that performance survives
distribution shift.

\section{Results and Discussion}
\label{sec:results}

To evaluate the proposed approach, we conducted extensive experiments to understand the robustness of the latent actions of ReLAP and whether it is translated to downstream policy performance. 
\begin{table}[t]
    \centering
    \caption{Performance comparison on LIBERO. 
    }
    \label{tab:literature-libero}
    \footnotesize
    \setlength{\tabcolsep}{4pt}
    \begin{tabular}{@{}lrrrrr@{}}
    \toprule
    Method & Spatial & Object & Goal & Long & Average \\
    \midrule
    \multicolumn{6}{l}{\textbf{Pretraining with physical action supervision}} \vspace{1mm} \\
    
    VLA-JEPA \cite{vlajepa} & 96.2 & 99.6 & 97.2 & 95.8 & 97.2 \\
    JALA \cite{jala} & 96.0 & 98.2 & 97.4 & 96.0 & 96.9 \\
    \midrule
    \multicolumn{6}{l}{\textbf{Action-free latent pretraining}} \vspace{1mm} \\
    LAPA \cite{lapa} & 73.8 & 74.6 & 58.8 & 55.4 & 65.7 \\
    UniVLA \cite{univla} & 96.5 & 96.8 & 95.6 & 92.0 & 95.2 \\
    ViPRA \cite{vipra} & --- & --- & --- & 79.0$^{\dagger}$ & --- \\
    RotVLA \cite{rotvla} & 98.2 & 99.6 & 98.4 & 96.4 & 98.2 \\
    ALAM \cite{alam} & 99.2 & 99.6 & 99.0 & 94.4 & 98.1 \\
    CoMo \cite{como} & 80.3 & 97.0 & 81.0 & 62.0 & 80.1 \\
    MoLA \cite{mola} & 93.0 & 99.5 & 99.5 & 96.0 & 97.0 \\
    \midrule
    ReLAP (ours) & \textbf{99.6} & \textbf{99.0} & 97.8 & 95.2 & 97.9 \\
    \bottomrule
    \end{tabular}
    \end{table}
    
    \begin{table*}[t]
    \centering
    \caption{Performance comparison on LIBERO plus.}
    \label{tab:literature-libero-plus}
    \footnotesize
    \begin{tabular}{@{}lrrrrrrrr@{}}
    \toprule
    Method & Cam. & Robot & Lang. & Light & Back. & Noise & Layout & Average \\
    \midrule
    \multicolumn{9}{l}{\textbf{Pretraining with physical action supervision}} \vspace{1mm} \\
    VLA-JEPA \cite{vlajepa} & 63.3 & 67.1 & 85.4 & 95.6 & 93.6 & 66.3 & 85.1 & 79.5 \\
    \midrule
    \multicolumn{9}{l}{\textbf{Action-free latent pretraining}} \vspace{1mm} \\
    UniVLA \cite{univla} & 1.8 & 46.2 & 69.6 & 69.0 & 81.0 & 21.2 & 31.9 & 43.9 \\
    LAWA \cite{lawa} & 69.2 & 66.7 & 62.8 & 96.2 & 64.5 & 85.5 & 78.6 & 74.4 \\
    \midrule
    ReLAP (ours) & 20.5 & 74.9 & 82.8 & 95.5 & 90.8 & 52.8 & 90.1 & 70.0 \\   
    \bottomrule
    \end{tabular}
    
    \end{table*}

\begin{table}[t]
\centering
\caption{Comparision of latent action probes of different latent action methods on LARY benchmark \cite{lary}}
\label{tab:lary}
\scriptsize
\setlength{\tabcolsep}{1.6pt}
\begin{tabular}{@{}lrrrrrrrr@{}}
\toprule
& \multicolumn{5}{c}{Regression MSE $\downarrow$} & \multicolumn{3}{c}{Classification acc.\ (\%) $\uparrow$} \\
\cmidrule(lr){2-6} \cmidrule(l){7-9}
Method & CALVIN & VLAB. & RoboC. & AgiBot & Avg. & Robot & Human & Avg. \\
\midrule
LAPA & 0.96 & 0.95 & 0.96 & 1.00 & 0.97 & 23.64 & 14.61 & 19.13 \\
UniVLA & 0.82 & 0.74 & 0.94 & 0.97 & 0.87 & 18.56 & 19.08 & 18.82 \\
villa-X & 0.86 & 0.72 & 0.94 & 0.97 & 0.87 & 29.90 & 17.80 & 23.85 \\
LAPA-DINOv3 & 0.50 & 0.25 & 0.82 & 0.84 & 0.63 & 27.04 & 64.19 & 45.62 \\
RotVLA & 0.30 & 0.08 & \textbf{0.35} & \textbf{0.07} & \textbf{0.20} & \textbf{67.62} & \textbf{74.33} & \textbf{70.98} \\
\midrule
ReLAP (ours) & \textbf{0.28} & \textbf{0.07} & 0.51 & 0.10 & 0.24 & 65.69 & 65.83 & 65.76 \\
\bottomrule
\end{tabular}
\end{table}

\subsection{Comparison with State-of-the-Art Methods}

Tables~\ref{tab:literature-libero}--\ref{tab:lary} compare ReLAP with recent latent-action models on standard LIBERO and LIBERO-Plus. Furthermore, we also compare the analysis of latent action representations on the LARY benchmark \cite{lary}. 

\noindent\textbf{LIBERO.}
ReLAP achieves \(97.9\%\) average success on LIBERO
(\(1{,}958/2{,}000\) episodes) 
 which is within \(0.3\) percentage points of RotVLA
(\(98.2\%\)) and \(0.2\) points of ALAM (\(98.1\%\)), while exceeding VLA-JEPA (\(97.2\%\)), MoLA (\(97.0\%\)), JALA (\(96.9\%\)), and
UniVLA (\(95.2\%\)). ReLAP obtains the highest spatial score among
the compared methods and remains competitive across the other suites.
The remaining failures are concentrated in LIBERO-Long, 
indicating that long-horizon execution
remains the primary limitation under the standard benchmark. It is worth mentioning that ReLAP uses \(11\)M pretraining samples which is substantially less than other competitive methods such as RotVLA, and ALAM. Furthermore, we use a \(2\)B-parameter model, of which only \(1.06\)B parameters are optimized compared to other large size models such as \(7\)B UniVLA. 

\noindent\textbf{LIBERO-Plus.}
Without any additional adaptation, we evaluate ReLAP zero-shot on
LIBERO-Plus and obtain a \(70.0\%\) macro-average success rate across
the seven perturbation categories. ReLAP is particularly robust to
robot initial-state (\(74.9\%\)) and object-layout (\(90.1\%\))
perturbations, achieving the highest reported performance in both
categories, while remaining competitive under language
(\(82.8\%\)), lighting (\(95.5\%\)), and background
(\(90.8\%\)) variations.
The primary robustness gap is concentrated in camera viewpoint and
sensor noise, where success drops to \(20.5\%\) and \(52.8\%\),
respectively. Consequently, ReLAP remains below the reported aggregate
performance of VLA-JEPA (\(79.5\%\)), and LAWA
(\(74.4\%\)), despite being competitive across most other perturbation
types. This limitation suggests that ReLAP is substantially more
robust to changes in task semantics and scene configuration than to
perturbations that directly alter the visual observation distribution.
We hypothesize that ReLAP's sensitivity to viewpoint and sensor
perturbations can be attributed by its Stage~2 pretraining setup,
where the VLM observes only the main camera view. Thus, although
Stage~3 introduces wrist-view information, the pretrained
vision-language representation has not been explicitly optimized for
cross-view consistency. The same qualitative weakness is visible in
UniVLA's LIBERO-Plus results, which also degrade sharply under camera
and sensor perturbations. In contrast, VLA-JEPA explicitly supports
multi-view visual conditioning during training, including two-view
world-model inputs, which may contribute to its stronger robustness
under camera variation. 

\noindent\textbf{LARYBench.}
Table~\ref{tab:lary} evaluates latent-action quality independent of
closed-loop policy performance. Under the LARY linear-probing protocol,
ReLAP achieves an average regression MSE of \(0.24\) and classification
accuracy of \(65.76\%\), second only to RotVLA (\(0.20\) and
\(70.98\%\), respectively). ReLAP achieves better regression error on
CALVIN (\(0.28\)) and VLABench (\(0.07\)). Its lower human-action
classification accuracy relative to RotVLA may partly reflect ReLAP's
pretraining mixture, which is dominated by robot-manipulation data.
The regression gap on RoboCOIN and AgiBotWorld should also be interpreted
in light of pretraining exposure: RotVLA includes both datasets in its
pretraining mixture, whereas ReLAP uses neither. These domains therefore
constitute genuine cross-dataset transfer tests for ReLAP. Despite this
distribution shift, ReLAP achieves MSEs of \(0.51\) on RoboCOIN and
\(0.10\) on AgiBotWorld, while remaining close to RotVLA in average
regression performance (\(0.24\) vs.\ \(0.20\)). Together, the LARY
results indicate that ReLAP learns compact, action-relevant transition
representations that generalize beyond its pretraining distribution.


\begin{figure*}[t]
\centering
\includegraphics[width=0.52\textwidth]{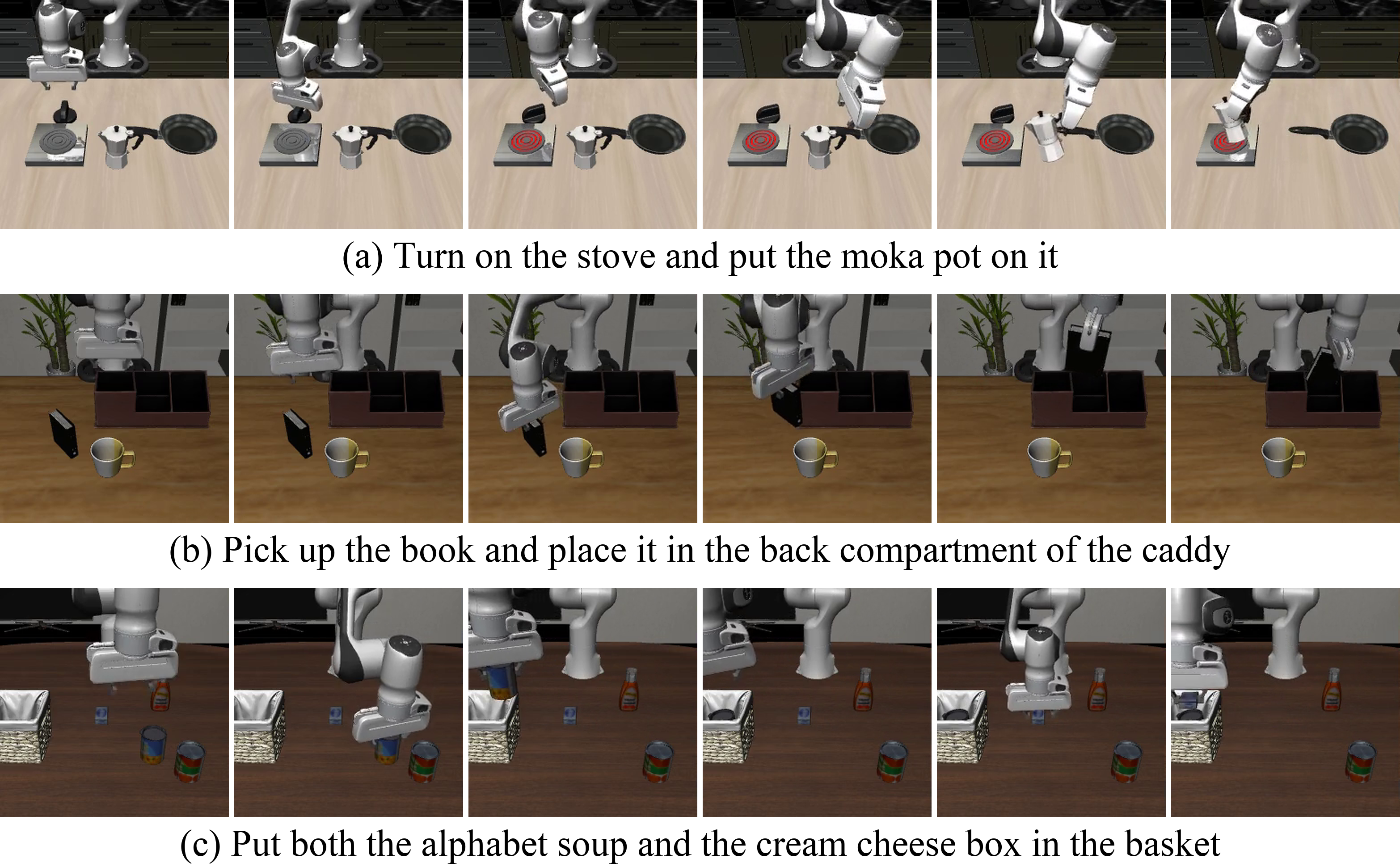}
\caption{Representative \method{} rollouts on three LIBERO-Long tasks, viewed from the main camera. Each row shows six frames of one successful episode in temporal order.}
\label{fig:rollouts}
\end{figure*}


\begin{table}[t]
\centering
\caption{latent Action Probe analysis of ReLAP}
\label{tab:latent_probes}
\footnotesize
\setlength{\tabcolsep}{3pt}
\renewcommand{\arraystretch}{1.08}
\begin{tabular}{@{}lrr@{}}
\toprule
Probe & Bridge (ID) & LIBERO-40 (OOD) \\
\midrule
\multicolumn{3}{@{}l}{\textbf{Dependence on the inferred latents} (prediction error, \%)} \\
Global mean latent & $+$34.80 & $+$28.65 \\
Zero latent & $+$77.70 & $+$99.67 \\
Swapping different/same visual transition & 0.439/0.074 & 0.471/0.211 \\
\midrule
\multicolumn{3}{@{}l}{\textbf{Temporal structure} (prediction error, \%)} \\
Sequence reversal & $+$4.03 & $+$2.50 \\
Single-slot corruption & $+$6.03--13.52 & $+$5.73--6.70 \\
\midrule
\multicolumn{3}{@{}l}{\textbf{Physical action Correspondance}} \\
Matched / mismatched action $R^2$  & 0.360/0.225 & 0.530/0.310 \\
\bottomrule
\end{tabular}
\end{table}

\subsection{Latent Action Representation Analysis}

A key challenge in latent action learning is to capture controllable, semantically meaningful transition dynamics from visual observations without merely encoding future-state information or action-irrelevant visual changes~\cite{zhang2025what}. We therefore analyze whether ReLAP latents are \emph{necessary for future prediction, sample-specific, temporally structured, and aligned with physical actions} as shown in Table \ref{tab:latent_probes}. We evaluate on \(1{,}536\) held-out Bridge windows as an in-distribution (ID) test set and \(2{,}936\) LIBERO-40 windows as out-of-distribution (OOD) transfer set unseen during Stage-1 training. 

\noindent\textbf{Latent dependency and sample specificity.}
We first test whether the recurrent world model genuinely depends on transition-specific latent information. We replace the inferred sequence with three controls: 
\emph{global mean}, which uses the dataset-average latent; and \emph{zero latent}, which removes latent conditioning entirely. These interventions increase latent world model prediction error by \(34.80\%\), and \(77.70\%\) on Bridge, and by \(28.65\%\), and \(99.67\%\) on LIBERO-40, respectively. Thus, accurate future prediction requires not merely a latent input, but the transition-specific information carried by the inferred sequence. We further test sample specificity by substituting a latent action from a different transition, and found that the latent world predictions degrade with latents from different transitions (0.439/0.074) on Bridge and (0.471/0.211) on LIBERO-40., highlighting the sample specific latent actions of ReLAP.




\noindent\textbf{Temporal Coherence}
We first test whether the four latent slots form a temporally coherent sequence. Reversing the latent order while preserving the same four latent vectors increases rollout error by \(4.03\%\) on Bridge and \(2.50\%\) on LIBERO-40. Corrupting a single latent slot causes a larger degradation of \(6.03\text{--}13.52\%\) and \(5.73\text{--}6.70\%\), respectively. These results indicate that the representation is temporally aligned and slot-specific: each latent captures transition information associated with a particular temporal interval, while global ordering has a smaller but measurable effect.

\noindent\textbf{Physical-action correspondence}
We next evaluate whether the temporal structure of the latent actions aligns
with robot actions. For each latent slot \(\mathbf{z}_i\), we linearly
predict each of the four future action segments. Temporally matched
latent--action pairs explain \(36.0\%\) of the action variance on Bridge,
compared with \(22.5\%\) for mismatched pairs, and \(53.0\%\) versus
\(31.0\%\) on LIBERO-40. Thus, each latent is preferentially predictive
of the physical action occurring during the same temporal interval,
providing direct evidence that the learned latent sequence is aligned
with the temporal structure of robot behavior.

\subsection{Qualitative Analysis of Results}




Across the examples, the policy successfully executes diverse multi-stage manipulation behaviors. In Fig.~\ref{fig:rollouts}(a), the robot first interacts with the stove and subsequently transports the moka pot to the target location, requiring sequential interaction with two distinct objects. Fig.~\ref{fig:rollouts}(b) demonstrates object-specific grasping and spatial placement of the book into the designated compartment of the caddy. Fig.~\ref{fig:rollouts}(c) further illustrates a longer compositional task in which two different objects must be sequentially manipulated and placed into the basket. These rollouts qualitatively demonstrate that the final VLM policy can ground language instructions into temporally coordinated, object-centric manipulation behaviors, including grasping, transport, placement, and multi-object sequencing. Notably, these behaviors are produced through direct VLM-to-action prediction at deployment, suggesting that the transition structure learned during latent-action pretraining can be transferred into the downstream policy without requiring the explicit latent-action pathway at inference.

\section{Discussion and Limitations}
\label{sec:discussion}

Our results also reveal two important limitations of the current ReLAP
design. First, recurrent latent rollout introduces an inherent temporal
asymmetry across the latent sequence. Earlier latent actions influence not
only their immediate transition, but also all subsequent predictions through
the recurrent rollout, and therefore receive stronger effective supervision
than later latent actions. This bias can favor accurate short-horizon dynamics
while making errors increasingly difficult to correct toward the end of the
\(1.6\,\mathrm{s}\) prediction horizon. 
Second, the current pretraining pipeline does not explicitly enforce
multi-view invariance. In Stage~2, latent prediction is learned primarily
from the main camera view, and no objective aligns representations across
different camera viewpoints. Although the Stage~3 action policy can access
the wrist image, the pretrained visual representation is therefore only
weakly constrained to remain consistent under viewpoint changes. 
Extending
the recurrent latent representation to substantially longer horizons and incorporating robustness to multiple views
remain important future
directions.
\section{Conclusion}

\method{} achieves superior performance on standard LIBERO with only 11M pretraining samples and 1.06B trainable parameters, remains competitive under zero-shot LIBERO-Plus perturbations, and learns latent representations that transfer strongly on the LARY benchmark. Our analyses show that these latents are sample-specific, temporally segmented, and aligned with future physical actions, including on datasets unseen during pretraining. These results suggest that action-free visual pretraining can provide compact and transferable supervision for downstream robot control without requiring an explicit latent-action bottleneck at deployment. The remaining gaps are concentrated in long-horizon manipulation and robustness to camera and sensor shifts, motivating more balanced long-horizon supervision and multi-view pretraining.


\bibliographystyle{IEEEtran}
\bibliography{references}

@misc{lapa,
  title = {Latent Action Pretraining from Videos},
  author = {Seonghyeon Ye and Joel Jang and Byeongguk Jeon and Sejune Joo and Jianwei Yang and Baolin Peng and Ajay Mandlekar and Reuben Tan and Yu-Wei Chao and Bill Yuchen Lin and Lars Liden and Kimin Lee and Jianfeng Gao and Luke Zettlemoyer and Dieter Fox and Minjoon Seo},
  year = {2025},
  howpublished = {arXiv:2410.11758},
  eprint = {2410.11758},
  archivePrefix = {arXiv},
  primaryClass = {cs.RO}
}

@misc{vipra,
  title = {{ViPRA}: Video Prediction for Robot Actions},
  author = {Sandeep Routray and Hengkai Pan and Unnat Jain and Shikhar Bahl and Deepak Pathak},
  year = {2026},
  howpublished = {arXiv:2511.07732},
  eprint = {2511.07732},
  archivePrefix = {arXiv},
  primaryClass = {cs.RO}
}

@misc{rotvla,
  title = {{RotVLA}: Rotational Latent Action for Vision-Language-Action Model},
  author = {Qiwei Li and Xicheng Gong and Xinghang Li and Peiyan Li and Quanyun Zhou and Hangjun Ye and Jiahuan Zhou and Yadong Mu},
  year = {2026},
  howpublished = {arXiv:2605.13403},
  eprint = {2605.13403},
  archivePrefix = {arXiv},
  primaryClass = {cs.RO}
}

@misc{jepalam,
  title = {Learning Latent Action World Models In The Wild},
  author = {Quentin Garrido and Tushar Nagarajan and Basile Terver and Nicolas Ballas and Yann LeCun and Michael Rabbat},
  year = {2026},
  howpublished = {arXiv:2601.05230},
  eprint = {2601.05230},
  archivePrefix = {arXiv},
  primaryClass = {cs.AI}
}

@misc{vjepa2,
  title = {{V-JEPA 2}: Self-Supervised Video Models Enable Understanding, Prediction and Planning},
  author = {Mido Assran and Adrien Bardes and David Fan and Quentin Garrido and Russell Howes and others},
  year = {2025},
  howpublished = {arXiv:2506.09985},
  eprint = {2506.09985},
  archivePrefix = {arXiv},
  primaryClass = {cs.AI}
}

@misc{vjepa21,
  title = {{V-JEPA 2.1}: Unlocking Dense Features in Video Self-Supervised Learning},
  author = {Lorenzo Mur-Labadia and Matthew Muckley and Amir Bar and Mido Assran and Koustuv Sinha and Mike Rabbat and Yann LeCun and Nicolas Ballas and Adrien Bardes},
  year = {2026},
  howpublished = {arXiv:2603.14482},
  eprint = {2603.14482},
  archivePrefix = {arXiv},
  primaryClass = {cs.CV}
}

@misc{lawm,
  title = {Latent Action Pretraining Through World Modeling},
  author = {Bahey Tharwat and Yara Nasser and Ali Abouzeid and Ian Reid},
  year = {2026},
  howpublished = {arXiv:2509.18428},
  eprint = {2509.18428},
  archivePrefix = {arXiv},
  primaryClass = {cs.RO}
}

@misc{
colaworld,
title={Co-Evolving Latent Action World Models},
author={Yucen Wang and Fengming Zhang and De-Chuan Zhan and Li Zhao and Kaixin Wang and Jiang Bian},
year={2026},
url={https://openreview.net/forum?id=WwwXi3rkUW}
}

@misc{wei2026learningadditivelycompositionallatent,
      title={Learning Additively Compositional Latent Actions for Embodied AI}, 
      author={Hangxing Wei and Xiaoyu Chen and Chuheng Zhang and Tim Pearce and Jianyu Chen and Alex Lamb and Li Zhao and Jiang Bian},
      year={2026},
      eprint={2604.03340},
      archivePrefix={arXiv},
      primaryClass={cs.CV},
      url={https://arxiv.org/abs/2604.03340}, 
}

@misc{univla,
  title = {{UniVLA}: Learning to Act Anywhere with Task-centric Latent Actions},
  author = {Qingwen Bu and Yanting Yang and Jisong Cai and Shenyuan Gao and Guanghui Ren and Maoqing Yao and Ping Luo and Hongyang Li},
  year = {2025},
  howpublished = {arXiv:2505.06111},
  eprint = {2505.06111},
  archivePrefix = {arXiv},
  primaryClass = {cs.RO}
}

@misc{libero,
  title = {{LIBERO}: Benchmarking Knowledge Transfer for Lifelong Robot Learning},
  author = {Bo Liu and Yifeng Zhu and Chongkai Gao and Yihao Feng and Qiang Liu and Yuke Zhu and Peter Stone},
  year = {2023},
  howpublished = {arXiv:2306.03310},
  eprint = {2306.03310},
  archivePrefix = {arXiv},
  primaryClass = {cs.AI}
}

@inproceedings{liberoplus,
  title = {{LIBERO-Plus}: A Progressive Robustness Benchmark for Visual-Language-Action Models},
  author = {Senyu Fei and Siyin Wang and Junhao Shi and Zihao Dai and Jikun Cai and Pengfang Qian and Li Ji and Xinzhe He and Shiduo Zhang and Zhaoye Fei and Jinlan Fu and Jingjing Gong and Xipeng Qiu},
  booktitle = {Proceedings of the IEEE/CVF Conference on Computer Vision and Pattern Recognition},
  pages = {38574--38583},
  year = {2026}
}

@misc{alam,
  title = {{ALAM}: Algebraically Consistent Latent Action Model for Vision-Language-Action Models},
  author = {Zuojin Tang and Haoyun Liu and Xinyuan Chang and Changjie Wu and Dongjie Huo and Yandan Yang and Bin Liu and Zhejia Cai and Feng Xiong and Mu Xu and Jiachen Luo and De Ma and Zhiheng Ma and Gang Pan},
  year = {2026},
  howpublished = {arXiv:2605.10819},
  eprint = {2605.10819},
  archivePrefix = {arXiv},
  primaryClass = {cs.RO}
}

@misc{dlam,
  title = {{DLAM}: Distributional Latent Actions with Temporal Constraints},
  author = {Zuojin Tang and Feifan Luo and Haoyun Liu and Botai Yuan and Dekang Qi and Ronghan Chen and Yandan Yang and Tong Lin and Xinyuan Chang and Mu Xu and Bin Liu and De Ma and Zhiheng Ma},
  year = {2026},
  howpublished = {arXiv:2607.27138},
  eprint = {2607.27138},
  archivePrefix = {arXiv},
  primaryClass = {cs.RO}
}

@misc{como,
    title = {{CoMo}: Learning Continuous Latent Motion from Internet Videos for Scalable Robot Learning},
    author = {Jiange Yang and Yansong Shi and Haoyi Zhu and Mingyu Liu and Kaijing Ma and Yating Wang and Gangshan Wu and Tong He and Limin Wang},
    year = {2025},
    howpublished = {arXiv:2505.17006},
  eprint = {2505.17006},
    archivePrefix = {arXiv},
    primaryClass = {cs.CV},
    note = {{CVPR} 2026},
  }

@misc{mola,
    title = {From Imagined Futures to Executable Actions: Mixture of Latent Actions for Robot Manipulation},
    author = {Yajie Li and Bozhou Zhang and Chun Gu and Zipei Ma and Jiahui Zhang and Jiankang Deng and Xiatian Zhu and Li Zhang},
    year = {2026},
    howpublished = {arXiv:2605.12167},
  eprint = {2605.12167},
    archivePrefix = {arXiv},
    primaryClass = {cs.RO},
    note = {{ICML} 2026},
  }

@misc{vlajepa,
    title = {{VLA-JEPA}: Enhancing Vision-Language-Action Model with Latent World Model},
    author = {Jingwen Sun and Wenyao Zhang and Zekun Qi and Shaojie Ren and Zezhi Liu and Hanxin Zhu and Guangzhong Sun and Xin Jin and Zhibo Chen},
    year = {2026},
    howpublished = {arXiv:2602.10098},
  eprint = {2602.10098},
    archivePrefix = {arXiv},
    primaryClass = {cs.RO},
  }

@misc{jala,
    title = {Joint-Aligned Latent Action: Towards Scalable {VLA} Pretraining in the Wild},
    author = {Hao Luo and Ye Wang and Wanpeng Zhang and Haoqi Yuan and Yicheng Feng and Haiweng Xu and Sipeng Zheng and Zongqing Lu},
    year = {2026},
    howpublished = {arXiv:2602.21736},
  eprint = {2602.21736},
    archivePrefix = {arXiv},
    primaryClass = {cs.RO},
    note = {{CVPR} 2026},
  }

@misc{lawa,
    title = {Latent Action as Intention Enables Efficient Future Imagination for World Action Models},
    author = {Xiang Li and Yupeng Zheng and Songen Gu and Huailiang Ma and Feng Yu and Yuhang Zheng and Xian Nie and Shanshuai Yuan and Yujie Zang and Weize Li and Shuai Tian and Moyang Liu and Ya-Qin Zhang and
    Wenchao Ding},
    year = {2026},
    howpublished = {arXiv:2608.24882},
  eprint = {2608.24882},
    archivePrefix = {arXiv},
    primaryClass = {cs.RO},
  }

@inproceedings{
zhang2025what,
title={What Do Latent Action Models Actually Learn?},
author={Chuheng Zhang and Tim Pearce and Pushi Zhang and Kaixin Wang and Xiaoyu Chen and Wei Shen and Li Zhao and Jiang Bian},
booktitle={The Thirty-ninth Annual Conference on Neural Information Processing Systems},
year={2025},
url={https://openreview.net/forum?id=DQMjemrVhe}
}

@misc{jepawam,
    title = {{JEPA-WAM}: Learning Vision-Language-Action Policies with Joint-Embedding World Modeling},
    author = {Yihan Lin and Jiawei He and Shifeng Bao and Chen Zhao and Yang Li and Xiaobo Wang and Yan Wang and Cheng Chi and Jing Zhang},
    year = {2026},
    howpublished = {arXiv:2608.09381},
  eprint = {2608.09381},
    archivePrefix = {arXiv},
    primaryClass = {cs.RO},
  }

@misc{lary,
  title = {{LARY}: A Latent Action Representation Yielding Benchmark for Generalizable Vision-to-Action Alignment},
  author = {Dujun Nie and Fengjiao Chen and Qi Lv and Jun Kuang and Xiaoyu Li and Xuezhi Cao and Xunliang Cai},
  year = {2026},
  howpublished = {arXiv:2604.11689},
  eprint = {2604.11689},
  archivePrefix = {arXiv},
  primaryClass = {cs.CV}
}

@misc{lapo,
  title = {Learning to Act without Actions},
  author = {Dominik Schmidt and Minqi Jiang},
  year = {2024},
  howpublished = {arXiv:2312.10812},
  eprint = {2312.10812},
  archivePrefix = {arXiv},
  primaryClass = {cs.LG},
  note = {{ICLR} 2024}
}

@misc{genie,
  title = {Genie: Generative Interactive Environments},
  author = {Jake Bruce and Michael Dennis and Ashley Edwards and Jack Parker-Holder and Yuge Shi and Edward Hughes and Matthew Lai and Aditi Mavalankar and Richie Steigerwald and Chris Apps and Yusuf Aytar and Sarah Bechtle and Feryal Behbahani and Stephanie Chan and Nicolas Heess and Lucy Gonzalez and Simon Osindero and Sherjil Ozair and Scott Reed and Jingwei Zhang and Konrad Zolna and Jeff Clune and Nando de Freitas and Satinder Singh and Tim Rockt{\"a}schel},
  year = {2024},
  howpublished = {arXiv:2402.15391},
  eprint = {2402.15391},
  archivePrefix = {arXiv},
  primaryClass = {cs.AI}
}

@misc{moto,
  title = {Moto: Latent Motion Token as the Bridging Language for Learning Robot Manipulation from Videos},
  author = {Yi Chen and Yuying Ge and Yizhuo Li and Yixiao Ge and Mingyu Ding and Ying Shan and Xihui Liu},
  year = {2024},
  howpublished = {arXiv:2412.04445},
  eprint = {2412.04445},
  archivePrefix = {arXiv},
  primaryClass = {cs.RO},
  note = {{ICCV} 2025}
}

@misc{claw,
  title = {{CLAW}: Learning Continuous Latent Action World Models via Adversarial Latent Regularization},
  author = {Tewodros Ayalew and Matthew Jeung and Samuel Wheeler and others},
  year = {2026},
  howpublished = {arXiv:2606.04130},
  eprint = {2606.04130},
  archivePrefix = {arXiv},
  primaryClass = {cs.LG}
}

@misc{flare,
  title = {{FLARE}: Robot Learning with Implicit World Modeling},
  author = {Ruijie Zheng and Jing Wang and Scott Reed and others},
  year = {2025},
  howpublished = {arXiv:2505.15659},
  eprint = {2505.15659},
  archivePrefix = {arXiv},
  primaryClass = {cs.RO}
}

@misc{dinowm,
  title = {{DINO-WM}: World Models on Pre-trained Visual Features enable Zero-shot Planning},
  author = {Gaoyue Zhou and Hengkai Pan and Yann LeCun and Lerrel Pinto},
  year = {2024},
  howpublished = {arXiv:2411.04983},
  eprint = {2411.04983},
  archivePrefix = {arXiv},
  primaryClass = {cs.RO}
}

@misc{oscar,
  title = {{OSCAR}: Omni-Embodiment Action-Conditioned World Model for Robotics},
  author = {Zhuoyuan Wu and Jun Gao},
  year = {2026},
  eprint = {2606.04463},
  archivePrefix = {arXiv},
  primaryClass = {cs.RO},
  howpublished = {arXiv:2606.04463},
  url = {https://arxiv.org/abs/2606.04463}
}

@misc{bridgev2,
  title = {{BridgeData V2}: A Dataset for Robot Learning at Scale},
  author = {Homer Walke and Kevin Black and Abraham Lee and Moo Jin Kim and Max Du and Chongyi Zheng and Tony Zhao and Philippe Hansen-Estruch and Quan Vuong and Andre He and Vivek Myers and Kuan Fang and Chelsea Finn and Sergey Levine},
  year = {2023},
  eprint = {2308.12952},
  archivePrefix = {arXiv},
  primaryClass = {cs.RO},
  howpublished = {arXiv:2308.12952},
  url = {https://arxiv.org/abs/2308.12952}
}

@misc{droid,
  title = {{DROID}: A Large-Scale In-The-Wild Robot Manipulation Dataset},
  author = {Alexander Khazatsky and Karl Pertsch and Suraj Nair and others},
  year = {2024},
  eprint = {2403.12945},
  archivePrefix = {arXiv},
  primaryClass = {cs.RO},
  howpublished = {arXiv:2403.12945},
  url = {https://arxiv.org/abs/2403.12945}
}

@misc{agibotworld,
  title = {{AgiBot World Colosseo}: A Large-scale Manipulation Platform for Scalable and Intelligent Embodied Systems},
  author = {{AgiBot-World-Contributors} and Qingwen Bu and Jisong Cai and others},
  year = {2025},
  eprint = {2503.06669},
  archivePrefix = {arXiv},
  primaryClass = {cs.RO},
  howpublished = {arXiv:2503.06669},
  url = {https://arxiv.org/abs/2503.06669}
}

@misc{egodex,
  title = {{EgoDex}: Learning Dexterous Manipulation from Large-Scale Egocentric Video},
  author = {Ryan Hoque and Peide Huang and David J. Yoon and Mouli Sivapurapu and Jian Zhang},
  year = {2025},
  eprint = {2505.11709},
  archivePrefix = {arXiv},
  primaryClass = {cs.CV},
  howpublished = {arXiv:2505.11709},
  note = {{ICLR} 2026},
  url = {https://arxiv.org/abs/2505.11709}
}

@misc{airoamoma,
  title = {{AIRoA MoMa} Dataset: A Large-Scale Hierarchical Dataset for Mobile Manipulation},
  author = {Ryosuke Takanami and Petr Khrapchenkov and Shu Morikuni and others},
  year = {2025},
  eprint = {2509.25032},
  archivePrefix = {arXiv},
  primaryClass = {cs.RO},
  howpublished = {arXiv:2509.25032},
  url = {https://arxiv.org/abs/2509.25032}
}

@misc{interndata,
  title = {{InternData-A1}: Pioneering High-Fidelity Synthetic Data for Pre-training Generalist Policy},
  author = {Yang Tian and Yuyin Yang and Yiman Xie and Zetao Cai and Xu Shi and Ning Gao and Hangxu Liu and Xuekun Jiang and Zherui Qiu and Feng Yuan and Yaping Li and Ping Wang and Junhao Cai and Jia Zeng and Hao Dong and Jiangmiao Pang},
  year = {2025},
  eprint = {2511.16651},
  archivePrefix = {arXiv},
  primaryClass = {cs.RO},
  howpublished = {arXiv:2511.16651},
  url = {https://arxiv.org/abs/2511.16651}
}

@misc{ego4d,
  title = {{Ego4D}: Around the World in 3,000 Hours of Egocentric Video},
  author = {Kristen Grauman and Andrew Westbury and Eugene Byrne and others},
  year = {2022},
  eprint = {2110.07058},
  archivePrefix = {arXiv},
  primaryClass = {cs.CV},
  howpublished = {arXiv:2110.07058},
  note = {{CVPR} 2022},
  url = {https://arxiv.org/abs/2110.07058}
}
\end{document}